\documentclass[letterpaper]{article} 
\usepackage{aaai24}  
\usepackage{times}  
\usepackage{helvet}  
\usepackage{courier}  
\usepackage[hyphens]{url}  
\usepackage{graphicx} 
\usepackage{natbib}  
\usepackage{caption} 
\usepackage{url}
\usepackage{amsmath,amssymb}
\usepackage{nameref}

\usepackage{algorithm}
\usepackage{algpseudocode}

\usepackage{xspace}
\usepackage{tikz}
\usepackage{pgfplots}
\usetikzlibrary{fit,positioning,shapes.misc}

\usepackage{changepage}
\usepackage{xcolor}

\newcommand{\fluke}{\textbf{\small \texttt{fluke}}\xspace}

\newcommand{\code}[1]{\mbox{
    \ttfamily
    \tikz \node[anchor=base,fill=blue!8]{\bfseries\scriptsize #1};
}}

\newcommand{\textttsmall}[1]{{\small\texttt{#1}}}

\definecolor{bgcli}{RGB}{60,60,100}
\newcommand{\flukecli}[1] {
  \vspace*{.3em}
  \begin{adjustwidth*}{}{+.2em}
  \begin{tikzpicture}
      \draw[rounded corners=3pt, draw=none, fill=bgcli] (0,0) rectangle (8.3,.6);
      \node[anchor=west] (text) at (0.1,0.25) {\scriptsize \texttt{\textcolor{white}{\$ #1}}};
      \node[opacity=.8] (console) at (7.7,0.47) {\tiny \textcolor{white}{\texttt{console}}};
  \end{tikzpicture}
  \end{adjustwidth*}
}

\newcommand{\flukeclis}[2] {
  \vspace*{.3em}
  \begin{adjustwidth*}{}{+.2em}
  \begin{tikzpicture}
      \draw[rounded corners=3pt, draw=none, fill=bgcli] (0,0) rectangle (8.3,1);
      \node[anchor=west] (text) at (0.1,0.6) {\scriptsize \texttt{\textcolor{white}{\$ #1}}};
      \node[anchor=west] (text) at (0.1,0.25) {\scriptsize \texttt{\textcolor{white}{\$ #2}}};
      \node[opacity=.8] (console) at (7.7,0.82) {\tiny \textcolor{white}{\texttt{console}}};
  \end{tikzpicture}
  \end{adjustwidth*}
}

\def\addlegendimage{\csname pgfplots@addlegendimage\endcsname}

\title{fluke: Federated Learning Utility frameworK for Experimentation and research}
\author{
    Mirko Polato
}
\affiliations{
    Department of Computer Science, University of Turin\\

    Corso Svizzera 185, 10149, Torino, Italy\\
    mirko.polato@unito.it
}

\begin{document}

\maketitle

\begin{abstract}
Since its inception in 2016, Federated Learning (FL) has been gaining tremendous popularity in the machine learning community. 
Several frameworks have been proposed to facilitate the development of FL algorithms, but researchers often resort to implementing
their algorithms from scratch, including all baselines and experiments. This is because existing frameworks are not flexible enough
to support their needs or the learning curve to extend them is too steep. In this paper, we present \fluke, a Python package designed 
to simplify the development of new FL algorithms. fluke is specifically designed for prototyping purposes and is meant for researchers
or practitioners focusing on the learning components of a federated system. fluke is open-source, and it can be either used out of the box
or extended with new algorithms with minimal overhead. 

\end{abstract}

\section{Introduction} \label{sec:intro}

Federated Learning (FL) is a machine learning paradigm where a model is trained across multiple
devices or servers holding local data samples, without exchanging them.
This paradigm has been gaining popularity in recent years due to its privacy-preserving properties,
as it allows training models on data that cannot be shared due to privacy concerns, such as medical
records, financial data, or user behavior.

The success of FL is evidenced by the constant increasing number of publications on the topic in top conferences. 

%
%

As a consequence of this growing interest, several FL frameworks
have been proposed, like Flower~\cite{flower}, TensorFlow Federated (TFF)~\cite{tff}, FATE~\cite{fate}, NVIDIA Flare~\cite{flare}, OpenFL~\cite{openfl}, just to name a few.
Although the cited framework are very powerful and flexible, they are not specifically designed for research purposes, and they are not easy to extend with new algorithms.
Flower is the only exception, as it is designed to be extensible and user-friendly, but still its configuration and customization are not trivial tasks\footnote{\url{https://flower.ai/docs/framework/tutorial-series-customize-the-client-pytorch.html}}.
Being not specifically designed for fast prototyping, these frameworks may seem somewhat over engineered.

It is common in FL research (ant in general) the need for quickly implement new algorithms to test new ideas, and this is not always easy with the mentioned frameworks. 
To support this claim, we can look at the number of recent papers proposing new FL algorithms with implementations not based on any framework, where all baselines and experiments have been implemented from scratch \cite{Dittorepo,FedExPrepo,FedNHrepo,FedNTDrepo,li2022federated,NIID-Bench,FedBNrepo,FedProtorepo,FedDynrepo,TurboSVMrepo}.
We cannot say for sure why these authors decided to re-implement everything from scratch, but we can speculate that the existing frameworks were not flexible enough to support their needs or the learning curve to extend them was too steep.

\fluke raises from the necessity of quickly prototyping new federated learning
algorithms in Python without worrying too much about aspects that are not specifically related to the algorithms themselves. \fluke also aims to improve the reproducibility of FL research, making it easier for researchers to replicate and validate studies.
\fluke is meant for researchers and practitioners focusing on the learning components of a federated system which is simulated on a single machine\footnote{One possible future development could be to allow \fluke to run on more machines.}.
In its current version (release 0.3.4), \fluke assumes a centralized architecture, where the simulated communication between the server and the clients is stable. 

We can summarize the main features of \fluke as follows:
\begin{itemize}
    \item \textbf{Open source}: \fluke is an open-source Python package. The code is written following the PEP8 guidelines and is comprehensively documented\footnote{\url{https://makgyver.github.io/fluke}}. 
    The code is also ``easy-to-read" because it is written to mimic the description and mathematical notation of the algorithm definitions as much as possible.
    The code is available on GitHub\footnote{\url{https://github.com/makgyver/fluke}}.
    \item \textbf{Easy to use}: \fluke is designed to be easy to use oof-the-shelf.
    Running a federated learning experiment is as simple as running a single command. 
    \item \textbf{Easy to extend}: \fluke is designed to be easy to extend minimazing the overhead of adding new algorithms. Implementing a new algorithm simply requires the definition of the client and/or the server classes.
    \item \textbf{Up-to-date}: \fluke comes with several state-of-the-art federated learning algorithms and datasets and it is regularly updated to include the latest affirmed\footnote{The algorithms are usually selected according to the venue and/or paper's citations.} techniques.
    \item \textbf{Simulated}: in \fluke the federation is simulated. This means that the communication between the clients and the server happens in a simulated channel and the data is not actually sent over the network. The simulated environment frees the user from aspects not related to the algorithm itself.
\end{itemize}

\fluke does not aim to be a full-featured FL framework, and it does not represent
a competitor to the existing ones mentioned above. Instead, it aims to be a utility framework for
researchers who need fast prototyping of new FL algorithms.
\section{\texttt{fluke}} \label{sec:fluke}

\fluke is available at \url{https://pypi.org/project/fluke-fl/} and can be installed using the following command:



\flukecli{\textbf{pip install} fluke-fl}

The installation allows to use both the \fluke Python API as well as the command line interface.
Being an open-source project available on GitHub at \url{https://github.com/makgyver/fluke}, it is also possible to start using \fluke by cloning the repository.


As mentioned in the introduction, \fluke simulates a federated learning environment and the
communication between the server and the clients. 
A very high level overview of how the simulation works is given in Algorithm~\ref{alg:fl}. The pseudo-code describes the main steps of a FL algorithm in \fluke. 
Clearly, specific algorithms may have different steps, but the general structure is the same.
\begin{algorithm}
\caption{Centralized Federated Learning in \fluke}\label{alg:fl}
\begin{algorithmic}[1]
    \Procedure{Server.fit}{}
    \State $\boldsymbol{\theta} \gets \text{init}()$ \Comment{Initialize the global model}
    \For{$t=1,2,\ldots,T$}\Comment{$T$ is the number of rounds}
        \State $E \gets $ select\_clients($C$)\Comment{$C$ is the set of clients}
        \State broadcast\_model($\boldsymbol{\theta}$, $E$)
        \For{$c \in E$}
        \State c.local\_training()
        \State $\boldsymbol{\theta}_c \gets$ receive\_model(c)
        \EndFor
        \State $\boldsymbol{\theta} \gets$ aggregate($\{\boldsymbol{\theta}_c\}_{c\in E}$)
    \EndFor
    \EndProcedure
    \State
    \Procedure{Client.local\_training}{}
	\State $\boldsymbol{\theta}_c \gets$ receive\_model(server)
    \State $\boldsymbol{\theta}_c \gets$ fit($\boldsymbol{\theta}_c$, $D_c$) \Comment{update $\boldsymbol{\theta}_c$ using the local data \label{alg:client-udate}}
    \State send\_model($\boldsymbol{\theta_c}$, server)
    \EndProcedure
\end{algorithmic}
\end{algorithm}
In \fluke the learing process is synchronous, i.e., the server waits for all the clients to finish their local training before aggregating the models.
The server is responsible for selecting the clients to participate in the training, broadcasting the global model to the clients, and aggregating the models received from the clients.
The clients are responsible for training the model on their local data and sending the updated model to the server.

\subsection{Architecture overview} \label{sec:fluke-architecture}

In this section, we provide an overview of the architecture of the \fluke package.
In Figure~\ref{fig:archi}, we show the main modules/submosules of \fluke and the interactions between them. 
The arrows mean that the source module uses the target module. The module \textttsmall{utils} is used by all the other modules.
In the following, we give a brief description of each \fluke's modules:

\begin{description}
    \item[data] contains the classes to load the datasets and to manage the data distribution (and training-test splitting) across the clients and the server. \fluke at the moment offers 13 datasets \cite{mnist,emnist,leaf,svhn,cinic10,cifar,tinyimagenet,fashionmnist} and five different non-IID distribution functions.
    \item[server] contains the base classes that define the server-side logic of a FL algorithm. The base server class behaves as in the FedAvg~\cite{fedavg} algorithm.
    \item[client] contains the base classes that define the client-side logic of a FL algorithm. The base client class behaves as in the FedAvg~\cite{fedavg} algorithm.
    \item[comm] contains the classes to be used for handling the communication between the server and the clients.
    Client and server \textit{must} exchange data (e.g., the model) or information through a (simulated) channel (\textttsmall{Channel} class). 
    Using the channel allows \fluke to keep track of the communication overhead (e.g., for logging purposes). Direct method calls between the parties are
    only used to trigger events (e.g., the server triggers the client-side local training).
    \item[algorithms] is the module where all the FL algorithms available in \fluke are defined.
    Each algorithm has its own submodule with all the classes needed to run the algorithm.
    Besides potential support classes/functions, each algorithm must define a class that inherit from \textttsmall{fluke.algorithms.CentralizedFL} and classes that inherit from \textttsmall{fluke.client.Client} and/or \textttsmall{fluke.server.Server}.
    \item[nets] is a support module with classes that define several neural networks used in the literature.
    \item[utils] is a module with a collection of utility functions and classes that are used across the package.
    \item[evaluation] contains classes to evaluate the performance of the algorithms.
\end{description}

\tikzstyle{s1} = [draw,minimum width=3.1cm,minimum height=.7cm]
\tikzstyle{s2} = [draw,minimum width=2.1cm,minimum height=.7cm]
\tikzstyle{s3} = [draw,minimum width=.6cm,minimum height=2.8cm]
\tikzstyle{s4} = [draw,minimum width=3.1cm,minimum height=.7cm]
\tikzstyle{s5} = [draw,minimum width=.6cm,minimum height=1.35cm]
\tikzstyle{s6} = [draw,minimum width=7cm,minimum height=.4cm]

\begin{figure}
    \centering
    \begin{tikzpicture}
        \node[s1,anchor=east,fill=red!80!black!50] (server) {\text{server}};
        \node[s2,below=.7 of server.south east,anchor=east,fill=yellow!20] (comm) {\text{comm}};
        \node[s3,left=.5 of server.south west, yshift=-.7cm,fill=green!60!black!60](algo) {\text{algo}};
        \node[s5,left=.4 of algo, yshift=.73cm,fill=lightgray!35]          (data) {\text{data}};
        \node[s5,below=.1 of data,fill=blue!30]                            (nets) {\text{nets}};
        \node[s3,right=.5 of server.south east, yshift=-.7cm,fill=orange!90!black!60]              (eval) {\text{eval}};
        \node[s4,below=.7 of comm.south east,anchor=east,fill=cyan!80!black!60]                  (client) {\text{client}};
        \node[s6,below=.1 of client,xshift=-.64cm]                         (utils) {\text{utils}};
        \node[above=0 of client.north west, xshift=.3cm, yshift=-.23cm] (clienttopleft) {};
        \node[above=0 of client.north east, xshift=-1.03cm, yshift=-.23cm] (clienttopright) {};
        \node[below=0 of server.south west, xshift=.3cm, yshift=.23cm] (serverbotleft) {};
        \node[above=0 of server.south east, xshift=-1.04cm, yshift=0cm] (serverbotright) {};
        \node[right=0 of algo.north east, xshift=-.22cm, yshift=-.35cm] (algotopright) {};
        \node[right=0 of algo.south east, xshift=-.22cm, yshift=.35cm] (algobotright) {};
        \node[right=0 of algo.north west, xshift=0cm, yshift=-.68cm] (algotopleft) {};
        \node[right=0 of algo.south west, xshift=0cm, yshift=.68cm] (algobotleft) {};
        \node[right=0 of eval.north west, xshift=0cm, yshift=-.35cm] (evaltopleft) {};
        \node[right=0 of eval.south west, xshift=0cm, yshift=.35cm] (evalbotleft) {};
        \node[above=0 of comm, xshift=-.5cm, yshift=-.23cm] (commtopleft) {};
        \node[above=0 of comm, xshift=.4cm, yshift=-.23cm] (commtopright) {};
        \node[below=0 of comm, xshift=-.5cm, yshift=.23cm] (commbotleft) {};
        \node[below=0 of comm, xshift=.4cm, yshift=.23cm] (commbotright) {};
        \draw[->,thick] (serverbotleft) -- (clienttopleft);
        \draw[->,thick] (serverbotright) -- (comm);
        \draw[->,thick] (clienttopright) -- (comm);
        \draw[->,thick] (data.east) -- (algotopleft);
        \draw[->,thick] (nets.east) -- (algobotleft);
        \draw[->,thick] (algotopright) -- (server.west);
        \draw[->,thick] (algobotright) -- (client.west);
        \draw[->,thick,dotted] (client.east) -- (evalbotleft);
        \draw[->,thick,dotted] (server.east) -- (evaltopleft);
    \end{tikzpicture}
    \caption{Overview of the \fluke's architecture with the interactions between the submodules.}
    \label{fig:archi}
    \end{figure}
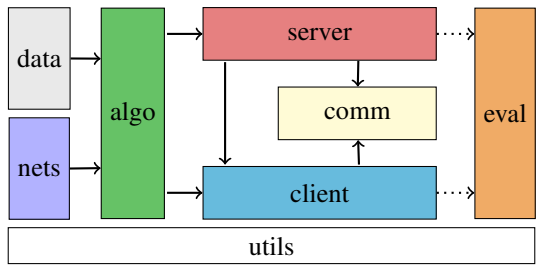

\tikzset{%
    baseline,
    inner sep=2pt,
    minimum height=10pt,
    rounded corners=2pt  
}

\subsection{\texttt{\textbf{fluke}} Command Line Interface (CLI)}
The \fluke CLI is the easiest and fastest way to run a federated learning experiment with \fluke.
It comes with two commands \code{fluke} and \code{fluke-get}.

\subsubsection{\textttsmall{fluke} command}

The \code{fluke} command is the main command of the \fluke package.
This command allows to run three different types of experiments (\code{EXP\_TYPE}):
\begin{itemize}
    \item Federated learning (\code{federation}): runs a centralized federated learning experiment;
    \item Centralized learning (\code{centralized}): runs a centralized training of a global model on the entire dataset;
    \item Clients only (\code{clients-only}): runs an experiment where each client independently trains a model on its own data.
\end{itemize}

Each experiment type requires two configuration files that must be passed as argument to the command.
Specifically, the \code{fluke} command has the following syntax\footnote{For simplicity we omit the potential \texttt{OPTIONS}. See the documentation for more details.}



\flukecli{\textbf{fluke} --config=EXP\_CFG EXP\_TYPE ALG\_CFG}

where:
\begin{itemize}
    \item \code{EXP\_CONFIG} is the path to the experiment configuration file (independent of the algorithm).
    In this file, the user can specify the dataset, the data distribution, the number of clients, the number of rounds, the seed, the device, and other parameters related to the experiment.
    \item \code{ALG\_CONFIG} is the path to the algorithm configuration file. Here, the user must specify the algorithm to use with all its hyper-parameters (e.g., the model, the server's and clients' hyper-parameters).
\end{itemize}

The division between the experiment and the algorithm configuration allows the user to run different algorithms on the same experimental setup. 

When the \code{fluke} command is used with the types \code{clients-only} and \code{centralized}, some of the parameters in the configuration files are not considered or not applicable, e.g, with \code{centralized}, the data distribution is ignored.
Instead, other parameters are computed accordingly, e.g., in the clients only case, the number of epochs is computed based on the number of rounds, the number of clients and the participation rate.

The output of this command is the evaluation of the algorithm on the test set(s) (according to the configuration) in terms of accuracy, precison, recall and F1 score (in both micro and macro average) at each round\footnote{The evaluation frequency can be set in the configuration file.}.

\subsubsection{\textttsmall{fluke-get} command}
The \code{fluke-get} command is a utility to get a configuration file template for the specified algorithm or for the experiment.
It directly downloads the template from the \fluke repository and saves it in the \textit{config} directory. The command has the following syntax:



\flukecli{\textbf{fluke-get} config CFG\_FILENAME}

where \code{CFG\_FILENAME} is the name of the configuration file (without the extension \textit{.yaml}) to download.
To get the list of available configuration files, the user can run the command \code{fluke-get list}.

\subsection{Documentation and tutorials}  \label{sec:fluke-doc}
In the spirit of making \fluke easy to use, we provide a comprehensive documentation that includes a quick start guide, and a developer guide.
The documentation is available at the following URL: \url{https://makgyver.github.io/fluke}, and it also includes a series of tutorials that show how to use \fluke in practice.
The tutorials on the API are structured as Python notebooks that can be run interactively on Google Colab.
\section{Use cases} \label{sec:use-cases}

\fluke can be employed in several scenarios related to FL research, however it is mainly designed for quick benchmarking and 
quick prototyping of new ideas. In this section, we address these two use cases showcasing how \fluke can be used in practice.

\subsection{Benchmarking} \label{sec:benchmarking}
Benchmarking is a common practice in machine learning research, and it is particularly important in federated learning, where the number of algorithms is increasing rapidly.
In this scenario, \fluke can be used off-the-shelf to compare the performance of different algorithms on the same dataset and under the same conditions.
To run a benchmark, one can use both the Python API and the command line interface (CLI) of \fluke.
However, the fastest way to run a benchmark of different FL algorithms is to use the \code{fluke} command with the \code{federation} experiment type.

Since the configuration of the experiment is decoupled from the algorithm, the user can run different algorithms on the same experimental setup very easily: the user can define a single experiment configuration file and then run different algorithms by changing the algorithm configuration, e.g.,


\flukeclis{\textbf{fluke} --config=exp.yaml federation fedavg.yaml}
{\textbf{fluke} --config=exp.yaml federation fedprox.yaml}

where \textit{exp.yaml} is the experiment configuration file and \textit{fedavg.yaml} and \textit{fedprox.yaml} are the two hypothetical configuration files for the FedAvg \cite{fedavg} and FedProx \cite{fedprox} algorithms, respectively.
If configured to log on \textit{Weights \& Biases} \cite{wandb}, the results of the different algorithms can be easily compared and analyzed.
Running a batch of experiments can be done by using a simple script that runs the \code{fluke} command with different algorithm configuration files.

The plot in Figure \ref{fig:fluke-plot} shows the accuracy of several algorithms run with \fluke using the same setting\footnote{Participation rate 20\%, learning rate 0.1 (no scheduling), batch size 32, 5 local steps.} with default hyper-parameters of the algorithms on the MNIST \cite{mnist} dataset.
Currently, \fluke comes with several state-of-the-art federated learning algorithms already implemented~\cite{fedavg,apfl,ccvr,ditto,fedamp,fedbn,fedbn2,fedbabu,fedavgm,feddyn,fedexp,fedlc,fednh,fednova,fedopt,fedper,fedproto,fedprox,pfedme,superfed,fedrep,lgfedavg,moon,scaffold,perfedavg,kafe2024}, and it is regularly updated to include the latest affirmed techniques.

\definecolor{color1}{HTML}{e41a1c}
\definecolor{color2}{HTML}{377eb8}
\definecolor{color3}{HTML}{4daf4a}
\definecolor{color4}{HTML}{984ea3}
\definecolor{color5}{HTML}{ff7f00}
\definecolor{color6}{HTML}{ffff33}
\definecolor{color7}{HTML}{a65628}
\definecolor{color8}{HTML}{f781bf}




\begin{figure}
\begin{tikzpicture}
\footnotesize
\begin{axis}[
	grid = both,
	major grid style = {lightgray},
	minor grid style = {lightgray!25},
	width = 0.46\textwidth,
	height = 0.25\textwidth,
	xlabel=round,
    xlabel near ticks,
	ylabel=accuracy,
	ylabel near ticks,
    legend pos=south east,
    legend cell align={left}
]
\addplot [thick,color1] table[col sep=comma, x=x, y=fedavg] {fluke-test.csv};
\addlegendentry{FedAVG}
\addplot [thick,color2] table[col sep=comma, x=x, y=scaffold] {fluke-test.csv};
\addlegendentry{SCAFFOLD}
\addplot [thick,color5] table[col sep=comma, x=x, y=fedexp] {fluke-test.csv};
\addlegendentry{FedExP}
\addplot [thick,color3] table[col sep=comma, x=x, y=feddyn] {fluke-test.csv};
\addlegendentry{FedDyn}
\addplot [thick,color4] table[col sep=comma, x=x, y=fedopt] {fluke-test.csv};
\addlegendentry{FedOpt}

\end{axis}
\end{tikzpicture}
\caption{Example of algorithms' performance with \fluke.} \label{fig:fluke-plot}
\end{figure}
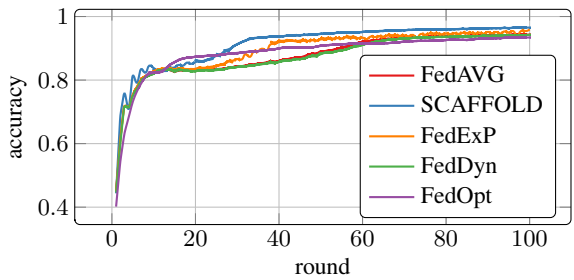

\subsection{Implementing a new algorithm} \label{sec:new-algorithm} \label{sec:fluke-new-alg}

The second main use case is the development of a new federated learning algorithm. 
In this scenario, the algorithm definition must be done using the \fluke Python API.
When a new federated learning algorithm is defined in the literature, its functioning is described by detailing the
behavior of the server and/or the clients. This is exactly how an algorithm is added in \fluke.

\textit{Note}: for space reasons, the following steps do not include all the necessary details, but they should be enough to understand the process. Please refer to the online documentation (\url{https://makgyver.github.io/fluke}) for more information, in particular the related tutorial \footnote{\url{https://makgyver.github.io/fluke/examples/tutorials/fluke_custom_alg.html}}.

\subsubsection{Defining the Client class}

In \fluke, the \textttsmall{Client} class represents the client-side logic of a FL algorithm. 
We suggest to start from this class when adding a new algorithm. 
A new client class \textttsmall{C} must inherit from \textttsmall{fluke.client.Client}\footnote{Or \texttt{PFLClient} in case of personalized FL.}.
The main methods that one may need to override are:

\begin{itemize}
    \item \textttsmall{fit}: this method is responsible for training the client's model on its data. This is usually the method that needs to be overridden. 
    \item \textttsmall{send\_model}: this method sends the model to the server. By default the model is meant to be the neural network weights, but it can be anything.
    The model must be sent through the channel (i.e., a \textttsmall{fluke.comm.Channel} object) that is an instance variable of the \textttsmall{Client} class.
    \item \textttsmall{receive\_model}: this method receives, through the channel, the model from the server.
    \item \textttsmall{finalize}: this method is called at the end of the learning (after the last round). It is useful to perform any final operations, e.g., fine-tuning.
\end{itemize}

\subsubsection{Definining the Server class}

The server-side logic of a FL algorithm is represented by the \textttsmall{Server} class in \fluke. A new server class \textttsmall{S} must inherit from \textttsmall{fluke.server.Server}.
The main methods that one may need to override are:
\begin{itemize}
    \item \textttsmall{fit}: this method is where all the federated training happens. Calling the server's fit method means starting the federation simulation. As long as the protocol follows the usual one in centralized FL, this method should not need to be overridden.  
    \item \textttsmall{aggregate}: this method is responsible for aggregating the models received from the clients. By default, it performs a (weighted) average of the client models.
    \item \textttsmall{broadcast\_model}: broadcasts the global model to the clients. By default, it is called at the beginning of each round.
    \item \textttsmall{finalize}: this method is called at the end of the learning (after the last round). Ideally, this method should be used to perform any final operation, e.g., to get the final evaluation of the model. In this method the server can trigger the finalization client-side.
\end{itemize}

\subsubsection{Defining the algorithm}

The final class to define is the one representing the whole algorithm, which must inherit from the \textttsmall{fluke.algorithms.CentralizedFL}\footnote{Or \texttt{fluke.algorithms.PersonalizedFL} in case of a personalized FL algorithm.} class.
In this class there is no logic involved, it is only the entry point of the algorithm and its sole purpose is to initialize the server and the clients.
The main methods that one should need to override are:

\begin{itemize}
    \item \textttsmall{get\_client\_class}: that returns the name of the class representing the client, if defined.
    \item \textttsmall{get\_server\_class}: that returns the name of the class representing the server, if defined.
\end{itemize}

After overriding these methods, the algorithm is ready to be run.

\subsubsection{Running the algorithm using the fluke CLI}

Once the algorithm is ready, it can be run using the \code{fluke} command.

Let's assume to have defined our algorithm, with all the necessary classes, in a python file named \textit{my\_algorithm.py} and that the class representing the algorithm is named \textttsmall{MyAlgorithm}.
The configuration file of the algorithm (e.g., \textit{alg\_cfg.yaml}) must contain in the key \code{name} the fully qualified name of the class, e.g., \code{name: my\_algorithm.MyAlgorithm}\footnote{Clearly, all potential hyper-parameters must be properly set in the configuration file.}.
Then, the algorithm can be run using the following command:


\flukecli{\textbf{fluke} ----config=exp.yaml federation alg\_cfg.yaml}

where \textit{exp.yaml} is the experiment configuration. The command must be run in the directory where \textit{my\_algorithm.py} is located.

\section{What's next?}\label{sec:future}
\fluke is a relatively young project, and there are several directions in which it could evolve without losing its original focus.
Our plan is to continue to integrate state-of-the-art algorithms, and to make \fluke more efficient.
Currently, \fluke is designed to run (mostly) single-threaded on a single machine, but in the future, we plan to support multi-threading and multi-device, for example leveraging the Ray\footnote{\url{https://www.ray.io/}} library.


\bibliographystyle{unsrt}
\bibliography{software,library}

\end{document}